\documentclass[letterpaper, 10 pt, conference]{ieeeconf}
\usepackage{graphicx}
\usepackage{psfrag}
\usepackage{pstool}
\usepackage{epsfig}
\usepackage{epstopdf}
\usepackage{float}
\usepackage{amsmath}
\usepackage{cite}
\usepackage{amsfonts}
\usepackage{amssymb}
\usepackage[top=0.75in, bottom=0.8in, left=0.75in, right=0.75in]{geometry}
\usepackage{nopageno}
\usepackage[]{algorithm2e}
\usepackage[utf8]{inputenc}
\usepackage{float}
\def\app#1#2{%
  \mathrel{%
    \setbox0=\hbox{$#1\sim$}%
    \setbox2=\hbox{%
      \rlap{\hbox{$#1\propto$}}%
      \lower1.1\ht0\box0%
    }%
    \raise0.25\ht2\box2%
  }%
}

\def\reals{ { {\rm  I \kern-0.15em R }  } }
\def\complex{\hbox{ \,{{\rm C} \kern-0.50em \raise0.20ex {  |}}\,}}

\def\lambdabf{\boldsymbol{\lambda}}

\def\pibf{\boldsymbol{\pi}}

\def\omegabf{\boldsymbol{\omega}}

\def\Deltabf{\hbox{\boldmath$\Delta$\unboldmath}}

\def\Dbf{\mathbf D}

\def\Mbf{\mathbf M}

\def\Qbf{\mathbf Q}

\def\Ubf{\mathbf U}

\def\abf{\mathbf a}
\def\bbf{\mathbf b}

\def\dbf{\mathbf d}

\def\fbf{\mathbf f}

\def\pbf{\mathbf p}
\def\qbf{\mathbf q}

\def\ubf{\mathbf u}
\def\vbf{\mathbf v}
\def\wbf{\mathbf w}
\def\xbf{\mathbf x}
\def\ybf{\mathbf y}

\def\0bf{\mathbf 0}
\def\1bf{\mathbf 1}

\def\be{\vskip .3cm \begin{equation}}
\def\ee{\end{equation} \vskip .4cm \noindent}

\newcounter{assumpnr}
\renewcommand{\theassumpnr}{\arabic{assumpnr}}

\newcommand{\ith}{\mbox{$i^{th}$}}

\renewcommand{\Pr}[1]{\text{Pr}\left\{ #1 \right\}}

\def\({\left(}
\def\){\right)}

\IEEEoverridecommandlockouts

\begin{document}
\title{Long-term Visual Localization using Semantically Segmented Images} %Labeled Point Feature Maps}

\author{Erik~Stenborg$^{1,2}$ Carl~Toft$^1$ and Lars~Hammarstrand$^1$% <-this % stops a space
\thanks{$^1$Chalmers University of Technology}% <-this % stops a space
\thanks{$^2$Zenuity}% <-this % stops a space
}

\maketitle

\thispagestyle{empty}

\begin{abstract}
Robust cross-seasonal localization is one of the major challenges in long-term visual navigation of autonomous vehicles. In this paper, we exploit recent advances in semantic segmentation of images, i.e., where each pixel is assigned a label related to the type of object it represents, to attack the problem of long-term visual localization. We show that semantically labeled 3D point maps of the environment, together with semantically segmented images, can be efficiently used for vehicle localization without the need for detailed feature descriptors (SIFT, SURF, etc.). Thus, instead of depending on hand-crafted feature descriptors, we rely on the training of an image segmenter. The resulting map takes up much less storage space compared to a traditional descriptor based map. A particle filter based semantic localization solution is compared to one based on SIFT-features, and even with large seasonal variations over the year we perform on par with the larger and more descriptive SIFT-features, and are able to localize with an error below 1~m most of the time.
\end{abstract}

\section{Introduction}
%Situation: tratta ner från autonom körning till lokalisering med kamera över längre tid
Although autonomous vehicle navigation can be done in uncharted environments, most efforts aiming at self-driving vehicles usable for every day activities, such as commuting, rely on pre-constructed maps to provide information about the road ahead. A central task for the self-driving vehicle is then to find its current location in these maps using observations from its on-board sensors, such as camera, lidar, radar etc. For this, in addition to navigational information, the maps typically describe the position of landmarks, i.e., points or structures in the environment, that can easily be detected by the on-board sensors. When it comes to cameras, it is common to use point features in the images as landmarks. The associated map is then constructed from these point features, where each feature is described by its 3D position in the world and a condensed description of the visual appearance of the local neighborhood around the feature. In the localization phase, these descriptors are used to find correspondences between point features in the current image and the features in the map \cite{fuentes2015visual, valgren2010sift}. A variety of methods for establishing these 2D-3D correspondences have been investigated, and once found, they can be used for calculating the full six-degrees-of-freedom camera pose \cite{Hartley2004, Li2012, Sattler2012}. 

% Why camera
The visual information captured by a camera is well suited for most tasks related to driving. If interpreted correctly, it can be used, e.g., to detect other road users or drivable road surface, and also for  self-localization. However, the abundance of information also provides several difficult challenges. 
For example, the appearance of feature points may change due to changes in light, weather, and seasonal variations. The traditional point descriptors used, e.g., SIFT, SURF, BRIEF, have been carefully designed to be robust towards uniform intensity changes and slight variation of viewpoint, but most were not designed to be invariant against large changes in lighting (day/night) or the fact that a tree looks completely different in summer compared to in winter. Additionally, it has been shown that the most commonly used feature detectors are very sensitive to changes in lighting conditions\cite{DTUinteresting}, implying that even if the feature descriptor is robust to these environmental changes, the resulting feature matches would still be incorrect since the detector does not trigger at the same points during localization as during mapping. Thus, when mapping and localization occur in sufficiently dissimilar conditions, it is very difficult to reliably match features between the image and the map, resulting in poor positioning accuracy or even complete failure of the localization algorithm. This long-term localization problem typically gets harder as the map gets older \cite{muhlfellner2016summary}, and is one of the major challenges in long-term autonomy.

\begin{figure}
	\centering
	\includegraphics[width=\linewidth]{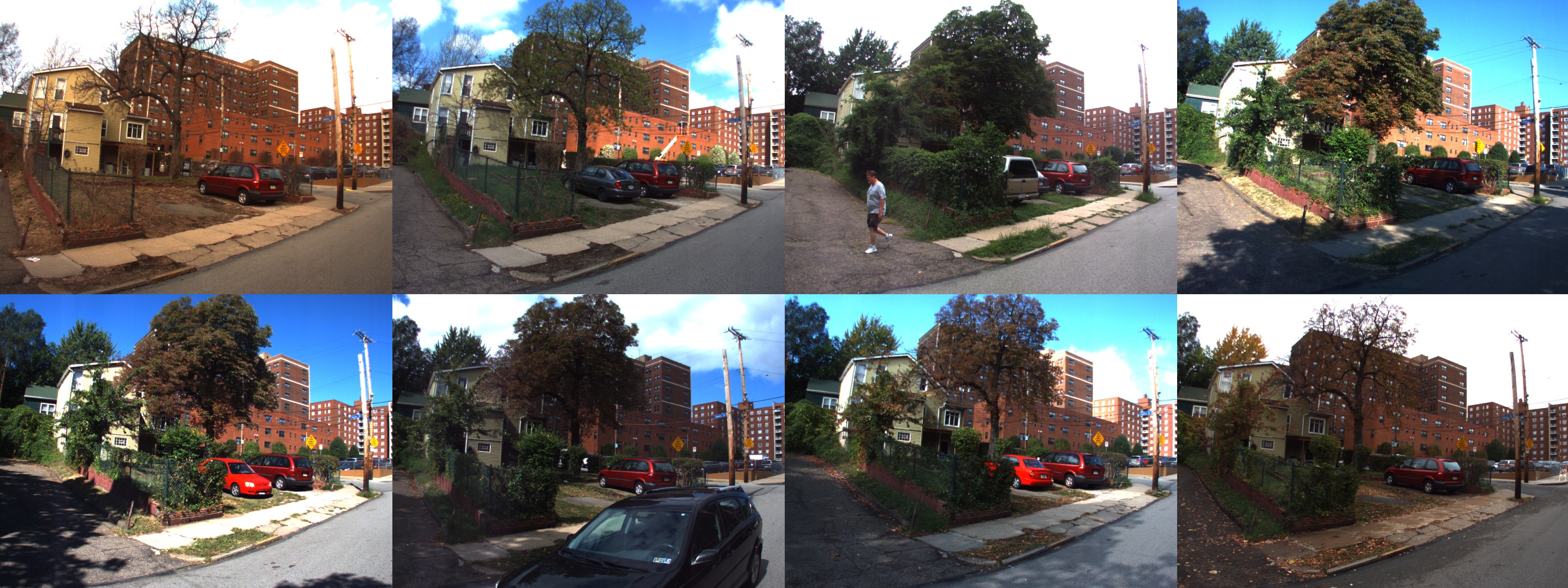}
	\includegraphics[width=\linewidth]{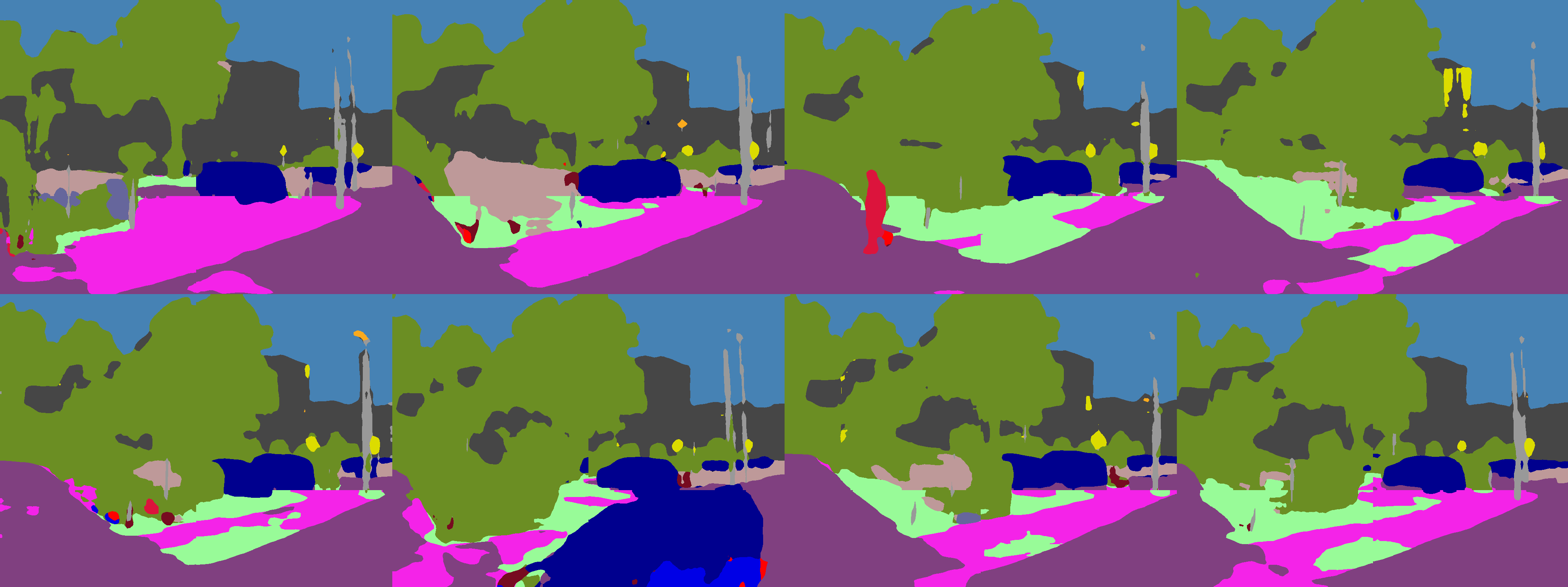}
	\caption{Example of how visual appearance changes with time for a scene (top half), and that semantic segmentation of the same images (bottom half) show less variation over time, although there are still large areas, especially around the tree and on the sidewalk which are misclassified. }
	\label{fig:works_seg}
\end{figure}

The problem can be boiled down to finding a description of the environment that is both usable for localization, invariant over time, and compact. If this can not be achieved, one has to cope with changing conditions by continuously updating the map \cite{muhlfellner2016summary, burki2016appearance, dymczyk2015gist}. An attempt at having more robust feature points and descriptors is presented in \cite{yi2016lift}, where they, instead of using handcrafted feature descriptors, train neural networks to produce more robust feature descriptors. Although they show promising results compared to SIFT they are not designed to handle the type of variations described above. 

% Taking a step back, matching of detected points in an image to points in a map is convenient if possible, but not strictly necessary for localization. If we relax this requirement, it is possible to make the descriptor less informative and possibly more robust. 

In this paper we propose to use recent advances in semantic segmentation of images \cite{Cordts2016Cityscapes}, and design a localization algorithm based on these semantically segmented images and a semantic point feature map, where, instead of using the traditional descriptors to describe our features, each point is only described by its 3-D position and the semantic class of the object on which it resides, similar to ideas presented in \cite{toft2017long}. By semantic class, we mean a classification into a few classes that are meaningful for a human, e.g. "road", "building", "vegetation", etc. The seasonal invariance is thus off-loaded from the feature descriptors to the semantic segmentation algorithm. %The map built this way is compact compared to the same point cloud with SIFT descriptors, and the time invariance problem is moved to the machine learning algorithm that does the semantic classification. 
The aim is to show that localization works well when the semantic classification is reasonably correct despite the more space efficient representation of the environment. The positioning performance of the proposed algorithm is compared to a localization algorithm based on a traditional SIFT point feature map, using data collected throughout a year.

%\section{Related works(?)}

\section{Problem statement}
This paper concerns the problem of sequentially finding the current position of a vehicle in a point feature map using on-board cameras. The dataset considered here comes from Carnegie Mellon University \cite{CMU2011}, and contains video, GPS measurements, and a "vehicle state" information which can be thought of as a rough truth signal or integrated odometry.
In this section we present the available observations in more detail and introduce notation for the map. We conclude by defining the problem at hand mathematically. 

\subsection{Observations}
\begin{figure}
\centering
\includegraphics[clip, trim=0cm 0.5cm 0cm 2.5cm, width=0.9\linewidth]{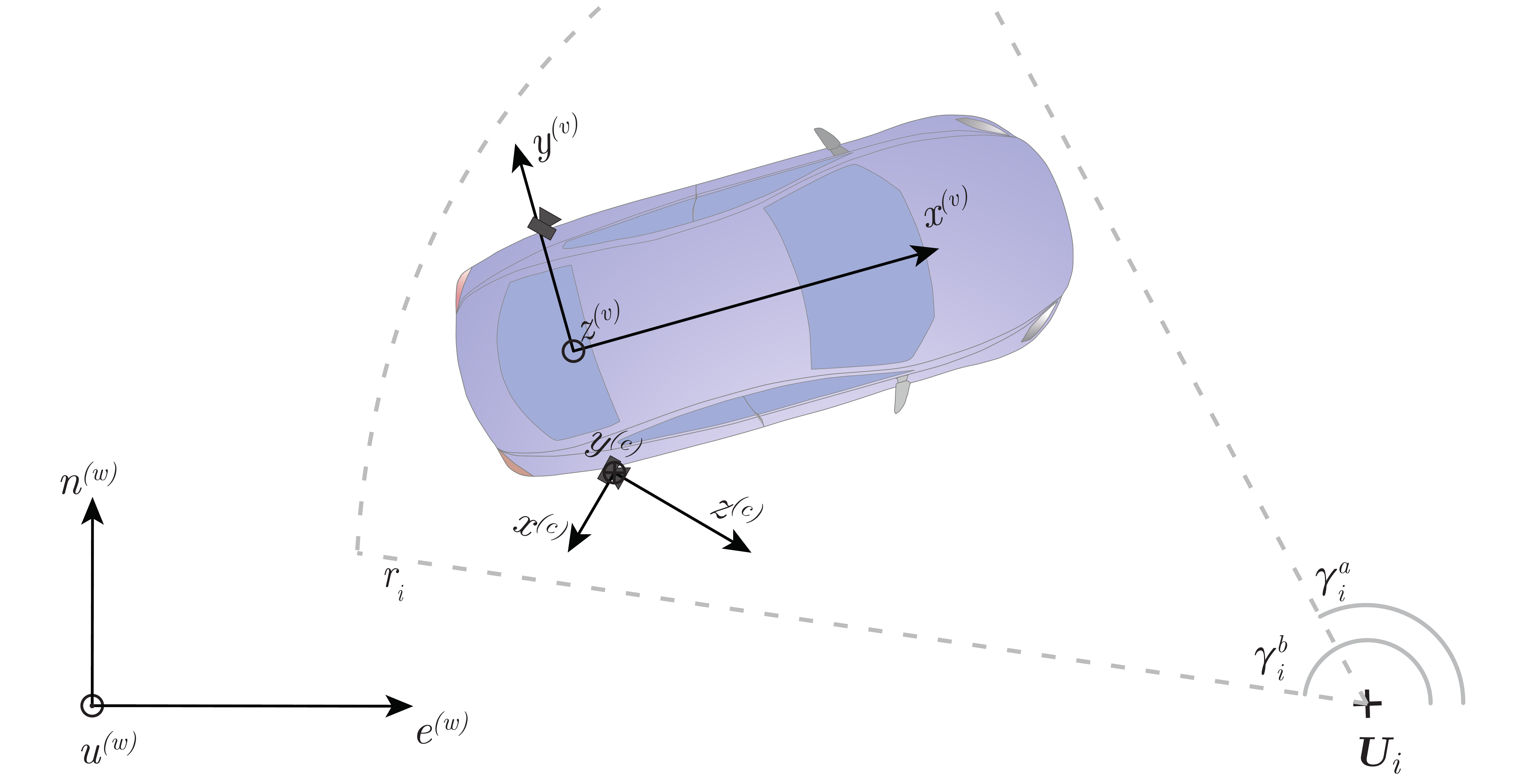}
\caption{Coordinate frames used are "world" in local ENU ($w$), vehicle ($v$), and camera ($c$). The origin of the vehicle frame is taken to be the mid point between the cameras, which gives us a horizontal lever arm, perpendicular to the direction of travel, to each camera. Also, the visibility wedge ($\gamma_i^a, \gamma_i^b, r_i$) of a map point, $\Ubf_i$, is illustrated.}
\label{fig:coordSys}
\end{figure}

The observations are taken with irregular, but known time intervals. We denote the time instance for which one such measurement was taken as $t$. 
Below follows a description of the information coming from the sensors at one of these time instances. Relevant coordinate frames and mounting positions are depicted in Fig. \ref{fig:coordSys}.

\subsubsection{Odometry}
From the vehicle state data it is possible to extract a 3-D velocity and 3-D rotational velocity, denoted $\vbf_t = [v^x_t,v^y_t,v^z_t]^T$ and $\omegabf_t = [\omega^z_t,\omega^y_t,\omega^x_t]^T$, respectively. The velocities are given relative the vehicle frame, and the superscripts indicate along or around which axis the component acts. All measurements are assumed to be affected by additive Gaussian noise, and the rotational velocity measurements are also affected by a slowly varying bias.

\subsubsection{Images}

The vehicle is equipped with a pair of calibrated cameras, mounted as indicated in Fig. \ref{fig:coordSys}. At a frequency of about 15 Hz each camera takes an RGB image with resolution 1024$\times$768. Although it is possible to use this raw image data directly \cite{PascoeICCV2015}, it is somewhat complicated. A more common approach is to condense the image into a set of feature points with associated descriptor vector, and view this as the measurement. As such, the image is pre-processed to produce a set of $n_t$ feature points and descriptor pairs, $\fbf_t = \{\langle\ubf^i_t, \dbf^i_t\rangle\}_{i=1}^{n_t}$, where $\ubf^i_t$ is a normalized image coordinate and $\dbf^i_t$ the associated descriptor vector.

In this paper, $\fbf_t$ will have different properties depending on which map we are using. In our proposed method (semantic point feature map) $\fbf_t$ will be dense and contain an element for each pixel in the image, see Fig. \ref{fig:works_seg}. In the case of a SIFT based map on the other hand, $\fbf_t$ is sparse and contains only the pixels for which the SIFT-algorithm has generated a detection and their associated SIFT-descriptors (a 128x1 vector).

\subsection{Maps}
We assume that we have a pre-constructed point feature map consisting of $M$ point features. Let us denote the map $\mathcal{M} = \{\langle\Ubf_i ,\, \Dbf_i, \, \mathcal{V}_i\rangle\}_{i=1}^M$. Each point feature is described by its global position $\Ubf_i=[U_i^e, U_i^n, U_i^u]^T$ (east, north and up, respectively), its associated descriptor vector, $\Dbf_i$ and visibility $\mathcal{V}_i = [\rho_i, \gamma_i^a, \gamma_i^b, r_i]^T$. 
The visibility of a feature point is parameterized by a probability of detection $\rho_i$ and a visibility volume defined by $\gamma_i^a, \gamma_i^b$, and $r_i$. The $\ith$ point is modeled to have a detection probability of $\rho_i$ in the wedge shaped volume defined by the two angles $\gamma_i^a$, and $\gamma_i^b$, in the horizontal plane, out to a range, $r_i$, from the point, and 0 elsewhere, see Fig. \ref{fig:coordSys}. 

\subsection{Problem definition}
The problem at hand is to recursively calculate the posterior density of the pose of the host vehicle relative to a map, $\mathcal{M}$, given all observations. That is, assuming that the pose of the host vehicle at time $t$ is described by the state $\xbf_t$, we want to sequentially calculate the density $p(\xbf_t|\fbf_{1:t}, \mathcal{M})$. Further, in this paper we assume that the vehicle state is given as $\xbf_t = [e_t, n_t, u_t, \gamma_t, \beta_t, \alpha_t]^T$, where $(e_t, n_t, u_t)$ is the position in the global coordinate frame and $(\gamma_t, \beta_t, \alpha_t)$ are the yaw, pitch and roll angles, respectively, of the vehicle in the same coordinate frame.

\section{Models}
For a filtering solution to the problem defined above, we need both a process model, describing how the state evolves over time, and measurement models that describe the relation between the state and our observations.

\subsection{Process model}
We model the vehicle using a simple point mass model. The process model consist of two parts. One part \eqref{eq:processModelPart1} models the motion by simply using the speed measurements as input, 
\begin{align}
\label{eq:processModelPart1}
\Mbf(\xbf_{t}) &= \Deltabf_{t} \Mbf(\xbf_{t-1}) \\
\Deltabf &= \begin{bmatrix}
e^{[\Delta t \omegabf_t + \qbf^\omega_t]_\times} & \Delta t \vbf_t + \qbf^v_t \\
\0bf & 1
\end{bmatrix}
\end{align} 
where $\Mbf(\cdot) \in SE(3)$ is the 4x4 matrix representation of a pose, $[\abf]_\times$ is the 3$\times$3 matrix such that $[\abf]_\times \bbf = \abf \times \bbf$ for all $\bbf$, $\Delta t$ is the time between the samples enumerated by $t-1$ and $t$, and the motion noise $\qbf^\omega_t \sim \mathcal{N}(\0bf, \Delta t \Qbf^{\omega})$, and $\qbf^v_t \sim \mathcal{N}(\0bf, \Delta t \Qbf^{v})$.

The other part of the process model is a term that ensures that there is always a little density left on the road. This would enable a lost filter to reacquire its lateral position even if it is lost. The road is defined by the route that the mapping vehicle drove, which is stored along with the 3-D landmarks in the map. By projection onto this trajectory, a small fraction of the density is moved to fall on the road. The complete process model can be described as a mixture,
\begin{align}
\label{eq:processModel}
p(\xbf_{t}|\xbf_{t-1}) = (1-\alpha) p_{m}(\xbf_{t}|\xbf_{t-1}) + \alpha p_{r}(\xbf_{t}|\xbf_{t-1}),
\end{align}
where $p_{m}(\cdot)$ is given by \eqref{eq:processModelPart1} and $p_{r}(\cdot)$ is the projection of $p_{m}(\cdot)$ onto the road.

\subsection{Measurement model}
\label{sec:meas_model}
We will present a measurement likelihood given the set of feature points in the current image $\fbf_t$, for both types of maps used in this paper, semantic and SIFT.

To arrive at a concise description of the likelihood we here assume that we know the correspondence between the points in the map and the points in the current image. As such, we have a data association vector $\lambdabf_t = [\lambda_t^1, \dots, \lambda_t^{n_t}]^T$, where $\lambda_t^i = j$ indicates that image feature $i$ corresponds to map feature $j$ if $j > 0$, otherwise the feature is not present in the map. Using this data association and assuming conditional independence between the pairs of $\ubf^i_t$ and $\dbf^i_t$, we get an expression for the likelihood as

\begin{align}\label{eq:generalLikelihood}
p(\fbf_t | \lambdabf_t, \xbf_t, \mathcal{M}) &= p(\{\langle\ubf^i_t, \dbf^i_t\rangle \}_{i=1}^{n_t}|\lambdabf_t, \xbf_t, \mathcal{M})\nonumber\\
&=\prod_i p(\ubf^i_t, \dbf^i_t|\lambdabf_t, \xbf_t, \mathcal{M})\nonumber\\
&=\prod_i p(\ubf^i_t, \dbf^i_t|\xbf_t, \mathcal{M}_{\lambda^i_t}).
\end{align}
where $\mathcal{M}_{\lambda^i_t}$ denotes the 3-D point with associated descriptor and visibility parameters in the map $\mathcal{M}$ which corresponds to feature $i$ according to the data association $\lambdabf_t$. So to be able to express \eqref{eq:generalLikelihood}, we need to define the model $p(\ubf^i_t, \dbf^i_t|\xbf_t, \mathcal{M}_{\lambda^i_t})$ for our two types of maps. How we handle the fact that $\lambdabf_t$ is not available from the measurements, is given in Section \ref{sec:AlgorithmicDetails}.

\subsubsection{SIFT map}
We start with a brief description of traditional localization in a point map with SIFT descriptors. With a given data association, the descriptor part of the feature will not contribute to the likelihood. We assume that the location of the SIFT detection in the image is subject to some noise, and a popular model for this is that the projection error of the 3-D points is zero mean and normally distributed, 
\begin{align} \label{eq:coordinateLikelihood}
p(\ubf^i_t, \dbf^i_t|  \xbf_t, \mathcal{M}_{\lambda^i_t}) & \propto
p(\ubf^i_t |  \xbf_t, \Ubf_{\lambda^i_t}) \nonumber \\
& = \mathcal{N}(\ubf^i_t; \pibf(\xbf_t, \Ubf_{\lambda^i_t}), \sigma_{\pi}^2),
\end{align}
where $\pibf(\cdot)$ is a standard pinhole camera projection model with lens distortion\cite{brown1971close}, and $\sigma_\pi^2$ is the variance of the detector error. Both the mounting of the camera relative the vehicle coordinate frame and its intrinsic parameters are implicit in the $\pibf(\cdot)$ function.

\subsubsection{Semantic map}

In the case of the semantic maps, both the descriptor of each map point, $D_j$, and image feature descriptor $d^i_t$ is a scalar class label from the Cityscapes classes \cite{Cordts2016Cityscapes}, i.e., $D_j \in \{\text{Building}, \text{Road}, \dots\}$. Further, as the semantic segmentation gives a class label for each pixel in the image, $\fbf_t$ is dense in the sense that it contains all the pixels in the image. 

Even though the descriptor for nearby pixels in the image clearly are correlated from the neural net classifier, we again make the simplifying assumption that the pixel class and pixel coordinates are independent and can thus partition the likelihood for a single feature point as
\begin{align}\label{eq:semPartitionedLikelihood}
p(\ubf^i_t, d^i_t&| \xbf_t, \mathcal{M}_{\lambda^i_t}) \nonumber\\
&=p(\ubf^i_t| \xbf_t, \Ubf_{\lambda^i_t})\Pr{d^i_t|  \xbf_t, \mathcal{M}_{\lambda^i_t}}.
\end{align}
This factorization would lead to overconfidence in the observations, and the it would get worse, the more measurements there are. This motivates the scale, $s$, and measurement cutoff, $N_c$, introduced in Section \ref{sec:semanticFilter}. The first factor, $p(\ubf^i_t| \xbf_t, \Ubf_{\lambda^i_t})$, denotes the probability of detecting a feature in pixel $i$. However, since all pixels in the input image are classified by the segmenter, and all pixels are used in the semantic model, pixel $i$ will always detect a feature and hence $p(\ubf^i_t| \xbf_t, \Ubf_{\lambda^i_t})$ is constant for all $i$. We thus obtain
\begin{align}\label{eq:semPartitionedLikelihood_2}
p(\ubf^i_t, d^i_t&| \xbf_t, \mathcal{M}_{\lambda^i_t}) \propto \Pr{d^i_t|  \xbf_t, \mathcal{M}_{\lambda^i_t}}.
\end{align}

Turning to the expression in the right hand side of \eqref{eq:semPartitionedLikelihood_2}, we have two cases: either there is no map point projected to this pixel, $\lambdabf_t^i = 0$, or there is one, $\lambdabf_t^i > 0$. In the first case, we have no information from the map about its class, and we assume a distribution for all such pixels which is simply the marginal distribution over all classes, %i.e. the probability that any pixel in the measured image will be a certain class,
\begin{align}
\Pr{d^i_t| \lambda_t^i = 0, \xbf_t, \mathcal{M}_{\lambda^i_t}} = \Pr{d^i_t}.
\label{eq:marginalPMF}
\end{align}

In the second case, the pixel coordinates corresponds to a point in the map but we are still uncertain if we detect the point or if it is occluded by something, e.g., a vehicle or pedestrian. To handle this uncertainty we introduce a detection variable $\delta$ which is 1 if the map point is detected in the image and 0 otherwise. Using this detection variable we can express the likelihood for the pixels with corresponding map points as 
\begin{align}
&\Pr{d^i_t| \xbf_t, \mathcal{M}_{\lambda^i_t}} = \nonumber\\
&\qquad= \sum_{\delta \in \{0,1\}}\Pr{d^i_t \big| \delta, D_{\lambda}^i} \Pr{\delta \big| \xbf_t, \mathcal{M}_{\lambda^i_t}}, 
\end{align}
where $\Pr{d^i_t \big| \delta = 0, D_{\lambda^i_k}}$ and $\Pr{d^i_t \big| \delta = 1, D_{\lambda^i_k}}$ are design PMF:s for the pixel class probability given that specific map point is occluded or visible, respectively. These are sensor specific models that also depend on the properties of the semantic segmentation algorithm used, and in section \ref{sec:mapCreation} we will discuss more how these are determined. The remaining model describes the probability that a given map point is visible and can be structured as
\begin{align}\label{eq:propDetected}
\Pr{\delta = 1\big| \xbf_t, \Ubf_{\lambda^i_k}, \mathcal{V}_{\lambda^i_k}}  &= v(\xbf_t, \Ubf_{\lambda^i_k},\mathcal{V}_{\lambda^i_k})\rho_{\lambda^i_k}(1-P_o), 
\end{align}
where $v(\cdot)$ is a function that is one if $\xbf_t$ is in the visibility wedge of the map point and $P_o$ is a design parameter specifying the probability that a visible map point is occluded. The probability for $\delta = 0$ is found as the reciprocal of \eqref{eq:propDetected}.

To conclude, the likelihood is factored into one part for the pixel coordinates $\ubf^i_k$ and one for the descriptor $d^i_k$. The first of the two factors is 1 for all pixels, and the second factor contributes to the product over all features in \eqref{eq:generalLikelihood} in two ways depending on whether or not there is a map point projected in the $i$:th pixel, according to 
\begin{align}
\label{eq:likelihood_func}
p(\fbf_t | \lambdabf_t, \xbf_t, \mathcal{M}) &= \prod_i p(\ubf^i_t, \dbf^i_t| \xbf_t, \mathcal{M}_{\lambda^i_t})\nonumber\\
&= \kern-.47cm \prod_{i \in \{i : \lambda^i_t > 0\}} \kern-.4cm p(d^i_t| \xbf_t, \mathcal{M}_{\lambda^i_t})\kern-.4cm\prod_{i \in \{i : \lambda^i_t=0\}} \kern-.4cm \Pr{d^i_t}.
\end{align}

\section{Algorithmic details}
\label{sec:AlgorithmicDetails}
\subsection{SIFT filter}
\begin{algorithm}[t]
	\label{alg:SIFT}
	\SetAlgoLined
	initialize $\hat{\xbf}_0$\\
	\For{each time instance $t$} {
		acquire image $\ybf_t$ \\
		motion update \eqref{eq:processModelPart1} \\
		extract SIFT points ($\ubf_t,\dbf_t$) from $\ybf_t$ \\
		$\pbf_t$ = projection of $\hat{\xbf}_t$ onto map trajectory \\
		select local map $\mathcal{M}_t$ from $\mathcal{M}$ using $\mathcal{V}$ and $\pbf_t$ \\
		match nearest neighbor $\dbf_t$ to $\mathcal{M}_t$\\
		RANSAC on $\ubf_t$ and $\Ubf$ from $\mathcal{M}_t$ to find $\lambdabf_t$\\
		\If{more than 7 inliers}{
			measurement update \eqref{eq:coordinateLikelihood}
		}
	}
	\caption{SIFT based localization}
\end{algorithm}

Now we have tractable models for both the motion and the two different classes of measurements based on SIFT descriptors and the semantic class descriptor. The measurement models are conditioned on a specific data association, and in the semantic class case, the model provides a simple way to make the correct data association, but for SIFT descriptor case we will describe the process further.

Our reference localization algorithm is similar to \cite{muhlfellner2016summary}, but instead of an iterative optimization we have implemented an Unscented Kalman Filter (UKF), and instead of iterative reweighting, we use RANSAC to select inliers from the proposal correspondences. The UKF makes use of the motion model \eqref{eq:processModelPart1} and measurement model for SIFT features \eqref{eq:coordinateLikelihood}, and is described in pseudo code in Algorithm \ref{alg:SIFT}. The "on road"-part of the process model \eqref{eq:processModel}, is in this case approximated by also including map points that are visible from the nearest point of the road, instead of only from the current estimate.

Before the SIFT detections can be used in \eqref{eq:coordinateLikelihood}, we must know the data associations, $\lambdabf_t$. Recall that $\lambdabf_t$ represents how point features are matched between the observed image and the map $\mathcal{M}$. We select one $\lambdabf_t$ by first selecting a subset of possibly visible points, $\mathcal{M}_t$, from the full map, $\mathcal{M}$, based on the current pose estimate and the visibility of the map points $\mathcal{V}$. Then the observed SIFT descriptors are matched in the image to their nearest neighbors ($L_2$-distance in the descriptor space) in this local map, $\mathcal{M}_t$, using Lowe's ratio criterion \cite{lowe2004distinctive} to select candidate matches. Then, to further cull false correspondences we employ a 3-point RANSAC approach in which three correspondence pairs are selected, and the four camera views that are consistent with these correspondences are calculated. Finally, we select the configuration giving the most inliers according to the reprojection error being less than a certain threshold, in our case 6 pixels.

\subsection{Semantic filter}
\label{sec:semanticFilter}
\begin{algorithm}[t]
	\label{alg:semantic}
	\SetAlgoLined
	initialize particles $\xbf_0$ and weights $\wbf_0$\\
	\For{each time instance $t$} {
		acquire image $\ybf_t$ \\
		motion update \eqref{eq:processModelPart1} \\
		project fraction of particles onto map trajectory \\
		assign class $\dbf^i_t$ to each pixel of $\ybf_t$ \\
		select local map $\mathcal{M}_t$ from $\mathcal{M}$ using $\mathcal{V}$ and $\xbf_t$\\
		measurement update \eqref{eq:weightUpdate}\\
		normalize weights $\wbf_t$, and resample if needed 
	}
	\caption{Sematic class based localization}
\end{algorithm}

For the localization using semantic data, we have chosen a bootstrap particle filter \cite{ristic2004beyond} to recursively estimate the posterior distribution as a sum of weighted Dirac delta functions. %A particle filter is in this case more suitable than, for example, a UKF as the likelihood is a probability mass function.

\begin{figure}[b]
	\centering
	\includegraphics[width=1\linewidth]{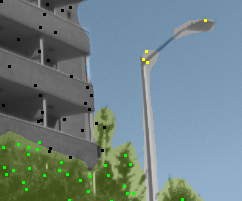}
	\caption{Cropped area from a segmented image with map points projected into it using the mean pose provided by the semantic localization filter. Mapping and localization are in this case separated in time by only 2 weeks. Black dots are map points which are classified mainly as building, green represents vegetation, and yellow is from the pole class.}
	\label{fig:imgCrop}
\end{figure}

To be able to evaluate the likelihood for a particle, we first need to determine which points in the map are potentially visible. This is similar to what is done for the SIFT case, and only needs to be done approximately and can thus be calculated for several nearby particles simultaneously, e.g. using their mean position together with the visibility parameters, $\mathcal{V}$, from the map. The potentially visible points are then projected to the image plane, creating a unique assignment, $\lambdabf_t$, from map to pixels for each particle. An illustration of map points projected into the segmented image is provided in Fig. \ref{fig:imgCrop}.

Dividing \eqref{eq:likelihood_func} by the constant ${\prod_{i}\Pr{d^i_t}}$ will simplify the weight update since we then only have to consider the pixels with a point projected into them,
\begin{align}
\label{eq:scaledLikelihood}
p(\fbf_t | \lambdabf_t, \xbf_t, \mathcal{M}) &\propto 
\frac{\prod_{i \in \{i : \lambda^i_t > 0\}} p(d^i_t| \xbf_t, \mathcal{M}_{\lambda^i_t})}{\prod_{i \in \{i : \lambda^i_t>0\}}\Pr{d^i_t}}.
\end{align}

Because we chose to model the measurements as conditionally independent when they are in fact not, the update we would get from this would be overly confident in the measurement, and to reduce this effect we raise the measurement likelihood to a positive number smaller than 1, so that the weight update for particle $j$ becomes
\begin{align}
\label{eq:weightUpdate}
w^{(j)}_{t} \leftarrow w^{(j)}_{t-1} \times p(\fbf_t | \lambdabf_t, \xbf^{(j)}_t, \mathcal{M})^{s/\max\{n_{\lambda_t},N_c\}}
%f(n) = \begin{cases}
%n, \quad & n>N_{c} \\
%N_{c}, \quad & n<= N_{c}
%\end{cases}
\end{align}
where $w^{(j)}_{t}$ is the weight associated with the $j$:th particle with state $\xbf^{(j)}_t$, $s$ is a scaling parameter set to 3, $n_{\lambda_t}$ is the number of map points that are projected in the image, and $N_c=400$ is a cutoff where more projected map points in the image do not contribute with more information, with the rationale that more points means their spacing in the image is smaller and thus their corresponding measurements are more correlated to each other.

The top level algorithm is summarized in pseudo code in Algorithm \ref{alg:semantic}.

%%Algo 2 was here

%\begin{enumerate}
%\item Roughly select from the map a subset of points that are probably visible using the prior information on position from previous time steps. [TODO: Define this more properly. Visibility function]
%\item Generate potential matches between the SIFT descriptors in the measurement and the subselected map points using nearest neighbor. 
%\item Discard matches that do not fulfill Lowe's ratio criterion to second best match\cite{lowe2004distinctive}. (For a probabilistic view on the ratio criterion \cite{kaplan2016interpreting}.)
%\item Use RANSAC with a 3 point minimal solver for pose to further discard matches that do not agree with the geometric constraints.
%\item The matches that survive are considered the true association between measurement and map if they are numerous enough.
%\item With the given data association, a measurement update is easily designed using the image coordinates and 3D-coordinates from the map, $p(\ubf_t|\xbf_t,\lambda_t,\mathcal{M}) = ...$
%\end{enumerate}

\section{Evaluation}
The localization framework is evaluated on the Carnegie Mellon University (CMU) visual localization dataset \cite{CMU2011, CMU2012}, where a test vehicle equipped with two cameras traversed a route of approximately nine kilometers in Pittsburgh sixteen times throughout the period September 1, 2010  - September 2, 2011. The route consists of a mix of urban and suburban areas, as well as green parks where mostly vegetation is visible in the cameras. We have used 12 of these 16 sequences in our evaluation. The selected runs capture the changes of the environment throughout the seasons, as well as a variety of weather and lighting conditions. 
The SIFT-features were extracted using VLFeat \cite{vedaldi08vlfeat}, and the semantic segmentation was done using Dilation 10 \cite{YuKoltun2016}.

\subsection{Map creation}
\label{sec:mapCreation}
For camera localization to work, we need a map to localize in. The focus of this paper is on the localization models, but we will give a small note on how we have chosen to handle the map, since there is no map included in the dataset. 
We have picked the first sequence of measurements from 1 September 2010 and created a map from that sequence of images. The remaining sequences are not used in the map creation, but only used to evaluate localization with respect to this map from Sep. 1. We used a structure-from-motion pipeline based on \cite{enqvist}, with the GPS and odometry constraints used to create an initial trajectory, after which a bundle adjustment procedure gave the final solution to landmarks and poses. Images at standstill and very low speeds were culled in order to avoid unnecessary computations. 
Because of limited computer resources, the whole sequence was split into smaller parts that were mapped separately. 
More on map creation and data processing for ground truth can be found in \cite{sattler2017benchmarking}.

\begin{figure}
	\centering
	\includegraphics[clip, trim=0cm 5cm 0cm 5cm, width=.7\linewidth]{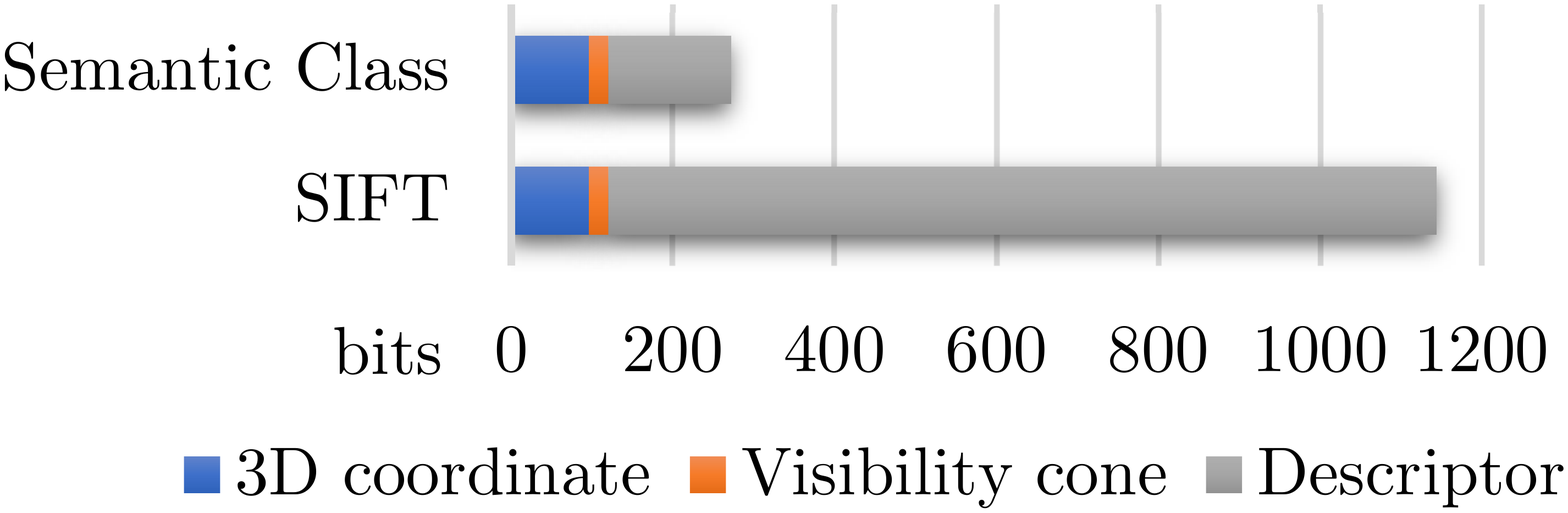}
	\caption{\label{fig:size} Storage space required for each point in the map. 3-D point and visibility cone are needed regardless of descriptor. The descriptors are quantized to 8 bit resolution. The semantic class descriptor could be even more compactly represented, since points rarely lie in corners where 3 or more classes meet, and can then be represented using only 1 or 2 bytes} 
	
\end{figure}

\begin{figure}
	\centering
	\includegraphics[clip, trim=0cm 3cm 0cm 3cm, width=\linewidth]{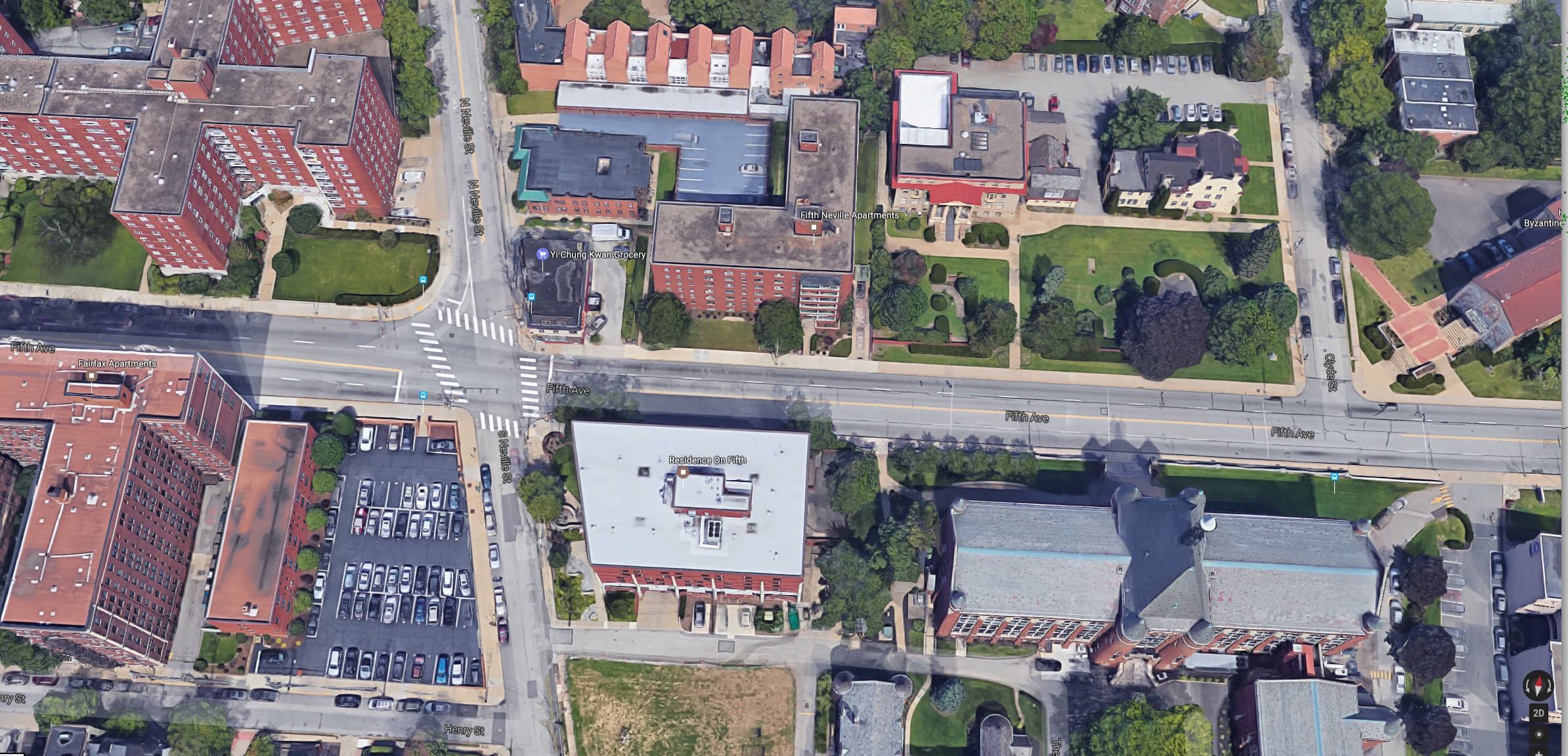}
	\includegraphics[width=\linewidth]{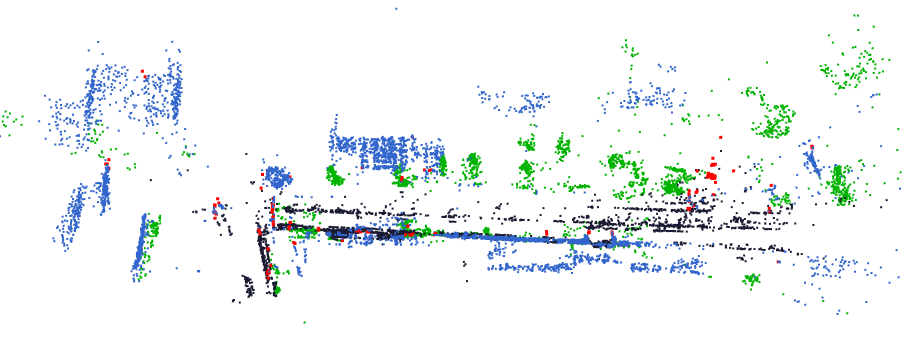}
	\caption{One part of the map, as viewed in Google Maps 3-D view (top), and the point cloud result of the structure from motion solution (bottom) colored by most likely category where blue is "construction", green is "nature", black is "flat" and red is "stationary object". }
	\label{fig:SemanticPointCloud_side}
\end{figure}

After calculating the 3-D points and camera poses, the descriptors for each point are determined. For the SIFT map, the arithmetic mean of the descriptors corresponding to each view of the 3-D point is taken as the descriptor of the 3-D point. For the semantic map, a small neighborhood of 7$\times$7 pixels around the detected point in each image is taken and then a normalized histogram over the classes of those pixels is used as the PMF $\Pr{d^i_t \big| \delta = 1, D_{\lambda^i_k}}$ directly. The marginal PMF from \eqref{eq:marginalPMF} is also calculated from data, as the normalized histogram for all pixels in all images in the mapping sequence. The last design PMF, $\Pr{d^i_t \big| \delta = 0, D_{\lambda^i_k}}$, is a manual adjustment of the marginal PMF. The dynamic objects, such as cars and pedestrians get increased probability while the stationary objects get decreased. We can see in fig. \ref{fig:size} that the size of SIFT descriptors is more than 6 times larger than even the most naive way of storing the semantic class descriptor. We have observed that normally each point only has probability mass in three or fewer classes, so if we encode only the top three most probable classes for each point, the descriptor size can be reduced to 39 bits. In fig. \ref{fig:SemanticPointCloud_side}, we show an example of the resulting point cloud from the map creation with semantic categories indicated by color.

\subsection{Ground truth}
The dataset provides some form of ground truth data in what is called "vehicle state", which includes pose, but it is not accurate enough to evaluate this type of localization. In order to obtain a more reliable ground truth, the sequences were aligned by adding manual correspondences between sequences and optimizing poses using the same structure from motion pipeline as in the map creation \cite{enqvist, sattler2017benchmarking}. The odometry measurements were generated from the relative motion between the ground truth poses, and then adding noise and bias. The added angular velocity bias was generated as $\bbf^{\omega}_t=(1-\gamma)\bbf^{\omega}_{t-1} + \qbf^{b}$ where $\qbf^{b} \sim \mathcal{N}(0,9\cdot10^{-10})$ and $\gamma = 10^{-5}$. Additional Gaussian noise with variance of $2.5\cdot10^{-5}$ was also added to the angular velocity, and noise with variance $4\cdot10^{-4}$ added to the velocity vector.

%\subsection{Parameters}
%The noise covariances of the process model \eqref{eq:processModel} are $\Qbf^{\omega}=2.5\cdot10^{-5}\Ibf_{3x3}$, and $\Qbf^{v}=10^{-2}\Ibf_{3x3}$ for angular velocity and velocity.
%From \eqref{eq:weightUpdate} the parameters are set to $N_c=400$ and $s=3$.

\section{Results}
\begin{figure}
\centering
\includegraphics[width=1\linewidth]{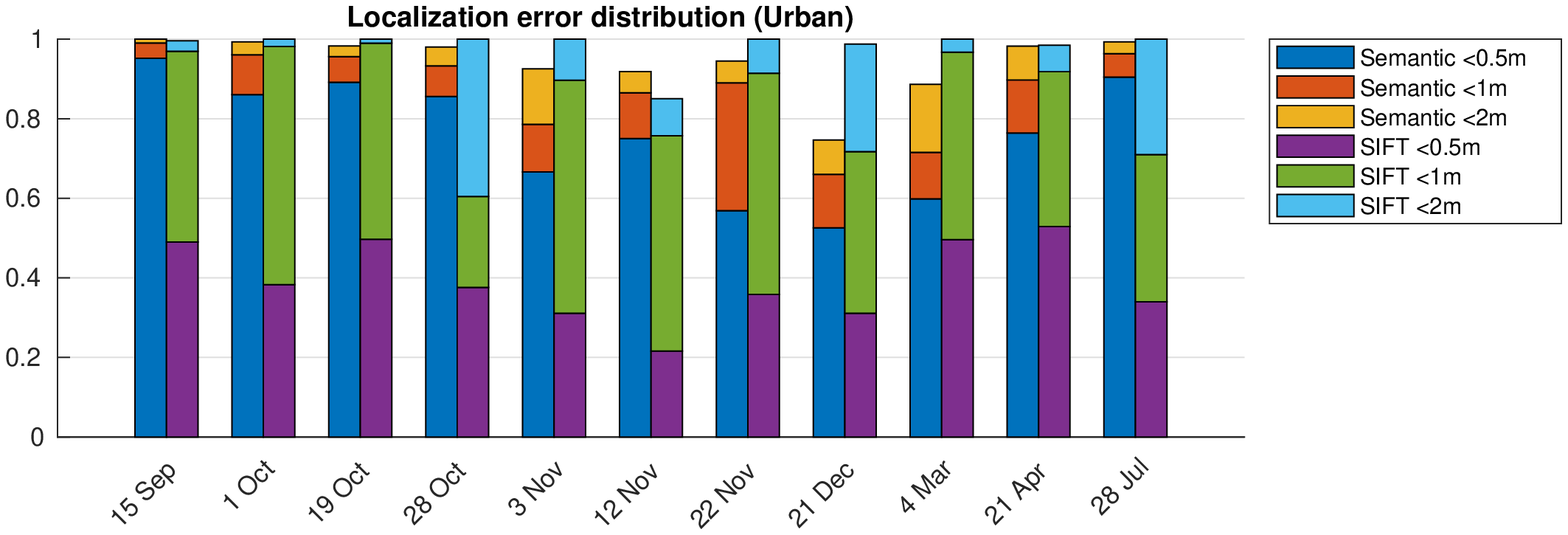}
\vspace{0.1pt}

\centering
\includegraphics[width=1\linewidth]{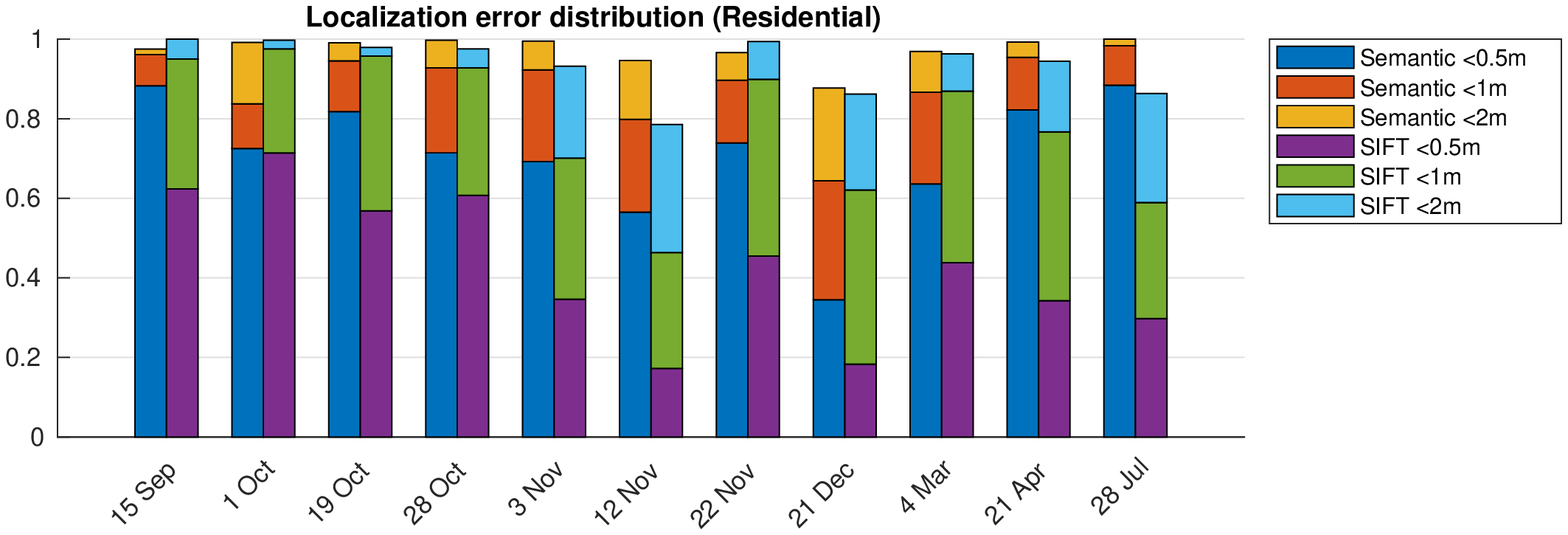}
\vspace{0.1pt}

\centering
\includegraphics[width=1\linewidth]{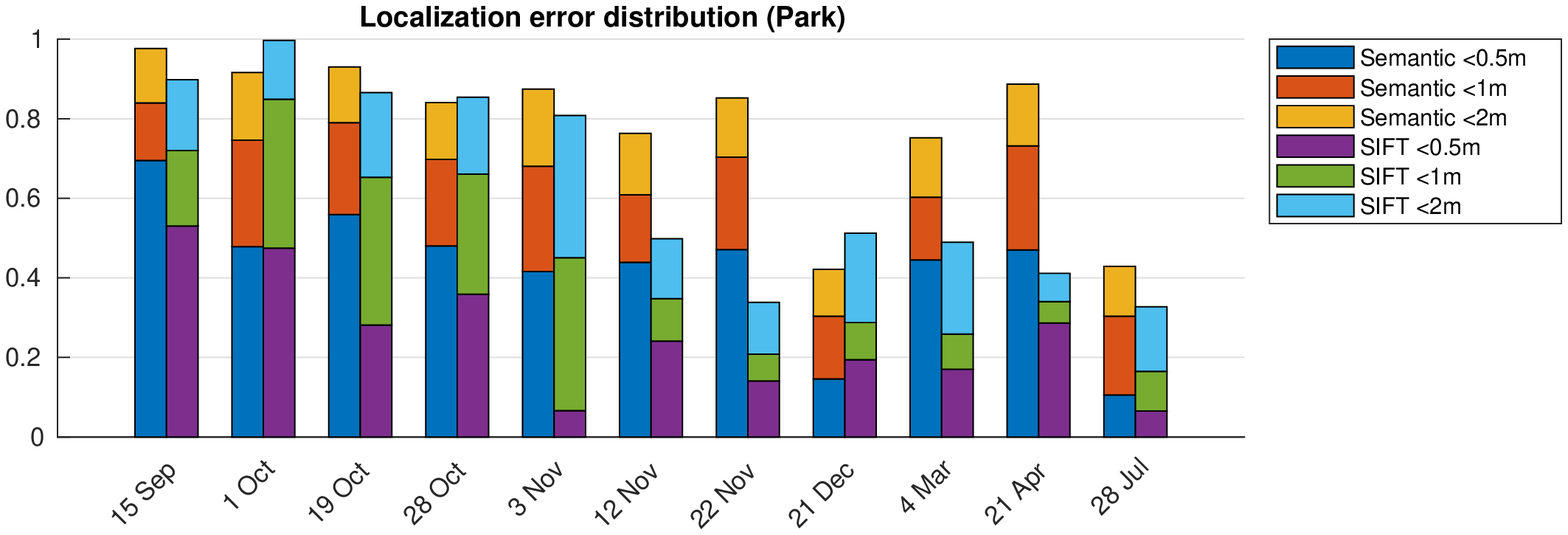}
	
\caption{\label{fig:errorHist} Histograms of the distance error of the three different environment types that the route was split in. Each histogram shows error distribution for the eleven runs and for both the semantic class based localization and SIFT based localization. Taller bars means that more samples were successfully localized within that limit, and are thus better. }
\end{figure}

%\begin{figure}
%\centering
%\includegraphics[width=.8\linewidth]{fig/error_rate.eps}
%\caption{\label{fig:error_rate} How often the error is below 1 m in the eleven localization runs for the semantic and SIFT filter.}
%\end{figure}

Overall semantic localization performed on par with the reference in our evaluation. Fig. \ref{fig:errorHist} shows the fraction of samples for which the localization error is within 0.5m, 1m, and 2m, for each of the 11 localization trials. 

\begin{figure}
\centering
\includegraphics[width=1\linewidth]{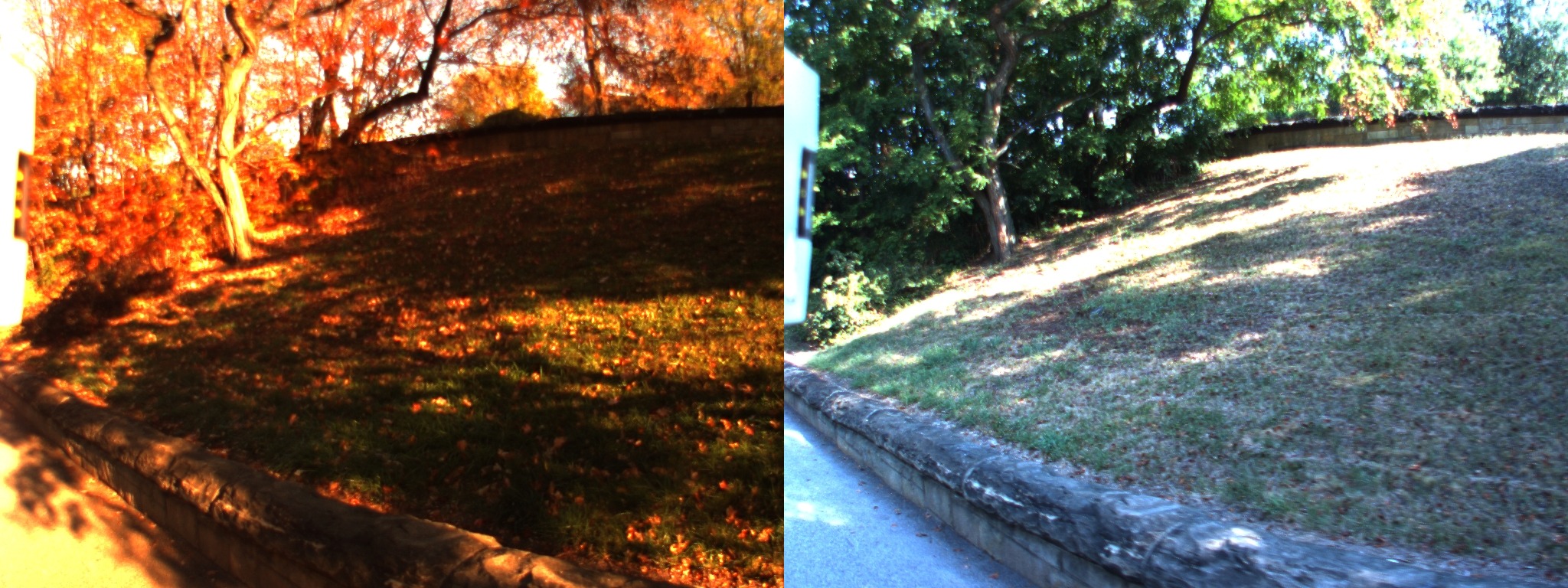}
\includegraphics[width=1\linewidth]{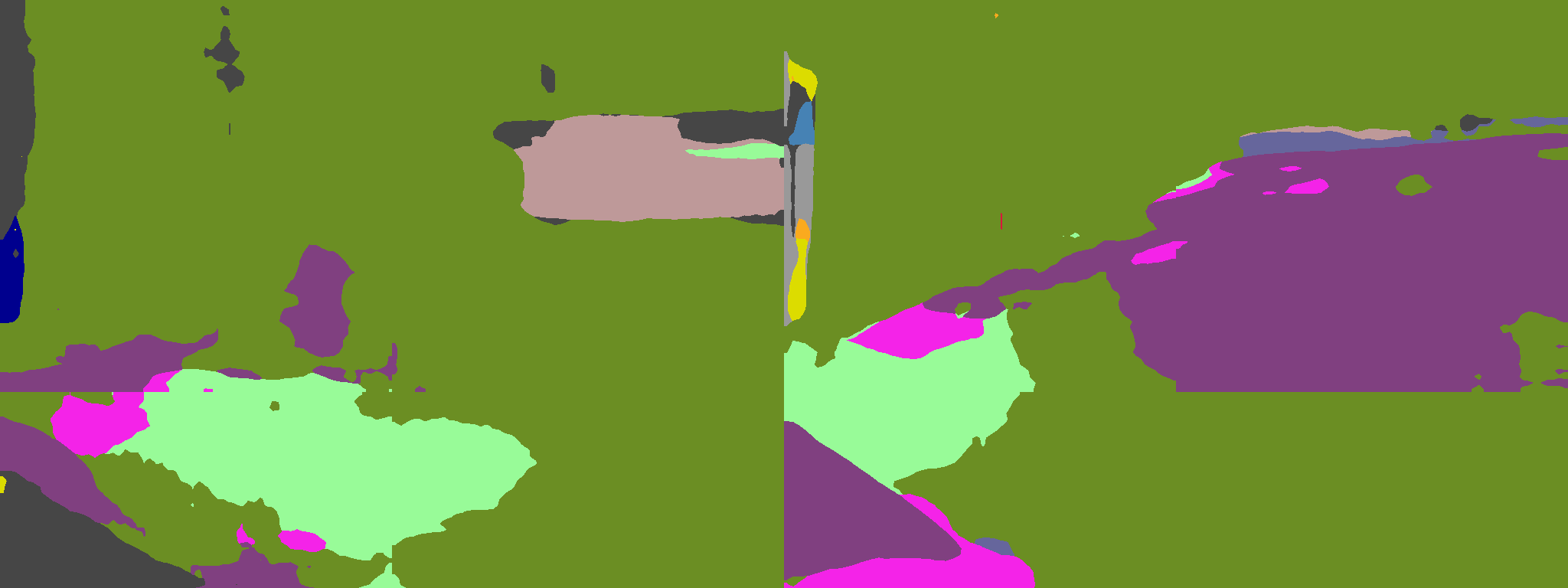}
\caption{Image and its segmentation for the place where semantic localization fails. Localization image to the left and mapping image to the right.}
\label{fig:fail_seg}
\end{figure}

When looking into a sample of cases where the localization error is large, we find two reasons for the semantic localization to fail. One is that the semantic class of the 3D points do not match the class in the image used for localization, because either or both are wrongly classified. E.g., the terrain beside the road is often misclassified as road  or vegetation (see Fig. \ref{fig:fail_seg}), and this seems to happen more frequently when the ground is covered in snow, as in the December sequence. 

Another reason for the semantic localization to fail, is the geometric configuration of certain scenes. On some roads which on the side have a monotonous class, e.g. a high wall, or a tall forest, the segmented images will look very similar regardless of where along the road the image is taken. In such scenes, the algorithm will give a higher likelihood score to a position where more points from the map happen to be visible. This could e.g. happen if there is a wall which has a lot of texture in one place, and which is almost uniform elsewhere. Then the localization filter adjusts the position such that the high texture place is visible for a longer time.

The SIFT localization typically fails when it is not able to find enough correspondences between the image and the map that survive the RANSAC step, and then the dead reckoning error accumulates. This was seen mostly in the "Park" area where virtually everything in the image is vegetation which drastically changes appearance over time.

%
%\begin{figure}
%\includegraphics[width=\linewidth]{localization_error_vehState3.pdf}
%\caption{\label{fig:localization_error} Localization error over time during evaluation of the filter on the dataset collected on October 1, 2010. }
%\end{figure}

\section{Discussion}

The results are promising in the sense that we can perform localization with results comparable to the reference algorithm, despite using much less informative point descriptors in the map. The results seem to not strongly support the goal of increased long term robustness, but from the first typical failure cases we have observed, we believe that results would improve if the segmentation algorithms were trained on data obtained during a larger range of environmental conditions (for example during winter, in more extreme lighting conditions and so on). The second problem of geometrically ambiguous configurations could possibly be helped by using other classes, e.g., adding road markings, splitting the vegetation class into trunk and foliage, etc. Without making these adjustments to the segmentation algorithm, the problematic scenarios that we set out to solve, are improved but not quite solved. Looking into adjusting the segmentation algorithm, and also investigating how to combine semantic localization with traditional feature point localization, would be interesting work in the future.

%We do not really count this as a failure of our localization algorithm, since this could probably be improved by training the segmentation algorithm on more data.

%\section{Acknowledgements}
%The first author has been funded in part by Vinnova and FFI, and the second author has been funded in part by the Swedish Research Council  (grant  no.  2016-04445), the Swedish Foundation for Strategic Research.

\medskip

\bibliographystyle{IEEEtran}
\bibliography{IEEEabrv,References}
% \renewcommand*{\bibfont}{\smaller}
% \printbibliography

\end{document}